\documentclass[a4paper, 10pt, conference]{article}      

\usepackage{graphics} 
\usepackage{mathptmx} 
\usepackage{times} 
\usepackage{amsmath} 
\usepackage{amssymb}  

\usepackage{microtype}

\usepackage[square,sort,comma,numbers]{natbib}
\usepackage{bibentry}
\nobibliography*
\usepackage{algorithm}
\usepackage{algpseudocode}
\usepackage{tikz}
\usepackage{pgfplotstable}
\usepackage{pgfplots}
\pgfplotsset{compat=newest}
\usepgfplotslibrary{groupplots}
\usepackage{tikzscale}
\usetikzlibrary{shapes.geometric, arrows,positioning,calc}
\title{\LARGE \bf
Simultaneous Task Allocation and Planning Under Uncertainty
}

\DeclareMathOperator*{\until}{\mathtt{U}}
\DeclareMathOperator*{\ltl_next}{\mathtt{X}}
\DeclareMathOperator*{\always}{\mathtt{G}}
\DeclareMathOperator*{\eventually}{\mathtt{F}}
\DeclareMathOperator*{\true}{\mathit{true}}
\DeclareMathOperator*{\safe}{\mathit{safe}}
\DeclareMathOperator*{\joint}{\mathit{joint}}

\newcommand{\A}{\mathcal{A}}
\newcommand{\M}{\mathcal{M}}
\newcommand{\G}{\mathcal{G}}

\author{Fatma Faruq$^{1}$, Bruno Lacerda$^{2}$, Nick Hawes$^{2}$ and David Parker$^{1}$
\thanks{$^{1}$School of Computer Science, University of Birmingham
        {\tt\small \{fxf603|d.a.parker\}@cs.bham.ac.uk}}%
\thanks{$^{2}$Oxford Robotics Institute, University of Oxford
        {\tt\small \{bruno|nickh\}@robots.ox.ac.uk}}%
}

\begin{document}
\begin{titlepage}
This paper is a preprint version of a paper accepted for publication at IROS 2018. 
\\
\\ 

Please cite this as
\\
\\

Faruq, F.; Lacerda, B.; Hawes, N. \& Parker, D.

Simultaneous Task Allocation and Planning Under Uncertainty

\emph{2018 IEEE/RSJ International Conference on Intelligent Robots and Systems,} 2018.

\end{titlepage}

\maketitle

\thispagestyle{empty}
\pagestyle{empty}

\begin{abstract}

We propose novel techniques for task allocation and planning
in multi-robot systems operating in uncertain environments.
Task allocation is performed simultaneously with planning,
which provides more detailed information about individual robot behaviour,
but also exploits independence between tasks to do so efficiently.
We use Markov decision processes to model robot behaviour
and linear temporal logic to specify tasks and safety constraints.
Building upon techniques and tools from formal verification,
we show how to generate a sequence of multi-robot policies, iteratively refining
them to reallocate tasks if individual robots fail,
and providing probabilistic guarantees on the performance
(and safe operation) of the team of robots under the resulting policy.
We implement our approach and evaluate it on a benchmark multi-robot example.

\end{abstract}

%
%
%
\section{INTRODUCTION}\label{sec:intro}

In many service robot applications, such as intra-logistics, surveillance or stock monitoring, it is desirable for a collection of \emph{tasks} to be allocated to a team of robots.
In this paper, we address applications such as these where tasks are independent (there are no inter-task dependencies) and each task only requires a single robot to complete it.
Most existing approaches for solving this class of problems divide the problem into separate \emph{task allocation} (TA) and \emph{planning} processes.
TA determines which robot should complete which tasks, and planning determines how each task, or conjunction of tasks, should be completed.
This separation is made to reduce the computational complexity of the problem.
It allows each robot to plan separately for its own task set, avoiding the need for a joint planning model which is typically exponential in the number of team members.

This separation also allows specialised algorithms to be used for the TA and planning parts, increasing the efficiency with which the task-directed behaviour of the team can be generated.
When doing this, TA usually assumes a greatly simplified model of planning in order to be able to efficiently compute allocations.
However, this separation also means that the TA process cannot be informed by the plans of the individual robots,
which prevents it from exploiting opportunities, or avoiding hindrances, that are only evident once planning has been performed.
For example, if the individual robots plan with time-based models, a task may be much quicker to complete at a particular time of day, but with TA separated from planning, this information cannot be exploited in the allocation process.

To address this limitation, recent work has considered the problem of \emph{simultaneous task allocation and planning} (STAP)~\cite{Schillinger2016,Schillinger2018ijrr}, which solves the complete problem in a single process, and can therefore take the plans of each robot (and their costs etc.) into account during the allocation process.
In this paper, we build upon the STAP approach and make the following contributions. 
We present the first formalisation of simultaneous task allocation and planning \emph{under uncertainty} (STAPU), and a method to solve this problem adapting techniques from formal verification of probabilistic systems.
%
We also contribute an extension of the approach to deal with reallocation of tasks when a robot fails.

Individual robots' capabilities and environments are described using Markov decision processes (MDPs).
The set of tasks to be completed by the team of robots is formally specified using linear temporal logic (LTL).
More precisely, tasks are defined in the  \emph{co-safe} fragment of LTL.
Furthermore a \emph{safe} LTL formula is provided to specify safety constraints to be obeyed by all robots.
Building upon techniques and tools for probabilistic model checking,
we propose methods to generate multi-robot policies that maximise the probability of successfully completing the set of tasks whilst satisfying the safety constraints.
We tackle the basic task allocation and planning problem using a \emph{team MDP}
that exploits the independence between tasks and adopts a sequential modelling approach to build policies for each robot.
We then iteratively improve these policies, by incorporating the possibility of reallocating
tasks in the event of individual robot failures.
To do so, we construct a precise joint model of the synchronised execution of the individual robot policies in order to identify states where robots might fail, and build new team policies from those states, thus providing \emph{probabilistic guarantees} on the performance (and safe operation) of the team of robots, along with an efficient task reallocation mechanism.
We implement our approach as an extension of the  probabilistic model checker PRISM~\cite{PRISM},
and evaluate its performance on a benchmark multi-robot example.

\section{RELATED WORK}


When looking at the existing literature in this area, we can consider the following distinctions: 1)
\emph{multi-agent path finding} (MAPF) approaches, which have rich models of inter-agent spatial interactions but can only solve path planning problems, 
versus more general planning approaches which can reason about a greater range of tasks; 
2) planning approaches which explicitly model uncertainty, versus those with deterministic models; 
 3) approaches which produce solutions using verification-based methods (and thus can produce guaranteed behaviour), versus other solution approaches; 
 and 4) approaches which integrate task allocation and planning, versus those that separate these processes. 


In robotics, MAPF~\cite{Felner2017} is a widely-studied problem which focuses on ensuring efficient, collision-free movement of a robot team through an environment. 
By focusing purely on path finding, domain-specific heuristics and algorithms can be used to efficiently solve larger problem instances than would otherwise be possible~\cite{sharon2015aij}.
A more general class of problems is multi-agent planning, which allows robot actions to have preconditions and effects across state variables in the problem, and can therefore represent a wider range of robot tasks~\cite{Zhang2016,Tumova2016a}. These typically focus on problems where interactions and coordination between agents are required to solve a single task (e.g. one robot needs to place an object onto a second robot), but these techniques do not usually involve explicit allocation of tasks.


A small number of approaches have looked into combining multi-agent TA and planning into a single problem~\cite{Schillinger2016,Schillinger2018ijrr,Claes,Ma2016}. 
Of particular relevance to this paper is the work described in~\cite{Schillinger2016} and~\cite{Schillinger2018ijrr} . They use a logical model of robot operation plus a team task specification in LTL and propose an algorithm that allocates tasks to robots in order to minimise the maximum cost any robot  will take to complete its tasks. Combining TA with planning allows TA to reason directly about how each robot can perform a task, but introduces the complexity of reasoning in a space which grows exponentially in the number of robots. To overcome this, the authors propose an approach where separate planning models for each robot are linked sequentially by \emph{switch transitions}, which allow one robot to pass tasks to the next robot in the team. They exploit these transitions to produce multi-robot plans which allocate tasks across the team in order to minimise the aforementioned solution metric of minimising the largest robot cost. This paper takes that work as a basis, but extends it to include uncertainty in the effects of robot actions. In  multi-robot systems, if uncertainty is not modelled, it is usually dealt with sub-optimally through execution monitoring and replanning.

The various planning fields surveyed above also have analogues which include uncertainty. MAPF approaches have included uncertainty to account for the performance of mobile robot localisation and navigation reliability~\cite{Wagner2017,Ma2017}.
Many single robot planning approaches assume that the environment is fully observable but responds probabilistically to robot actions and thus formulate planning problems using MDPs~\cite{Lahijanian2016,Smith}. Approaches in this space include our prior work~\cite{Lacerda2015a,Lacerda2014} which uses verification-based methods to produce probabilistically-guaranteed behaviour policies for a mobile robot, where elements of the MDP are learnt from experience. 
%
When extending MDP planning approaches to multi-robot settings, authors either assume communication and sparse interactions between robots in order to maintain full observability and mitigate scalability issues~\cite{Scharpff2016a,Claes:2015:EAM:2772879.2773265}; resort to auctioning approaches for TA, thus keeping the planning over single robot models~\cite{spaan2010icra,capitan2013ijrr,schillinger2018rss};  or otherwise use the computationally-demanding decentralised, partially-observable MDP (DecPOMDP) formalisation which accounts for the unknown state of other robot in the problem~\cite{Amato2015}.
As is appropriate in many service robot domains, we make the assumption of perfect communication, thus allowing this work to retain the MDP formalisation.

\section{PRELIMINARIES}

\subsection{Markov Decision Processes}

We use \emph{Markov decision processes} (MDPs) to model the evolution of robots and their environment. 
An MDP is a tuple $\mathcal{M}=\langle S, \overline{s},A,\delta_\mathcal{M},AP,Lab\rangle$, where:  $S$ is a finite set of states;
$\overline{s} \in S$ is the initial state;
$A$ is a finite set of actions;
$\delta_\mathcal{M}:S \times A \times S \rightarrow [0,1]$ is a probabilistic transition function, where $\smash{\sum_{s' \in S} \delta_\mathcal{M}(s,a,s') \in \{0,1\}}$ for all $s, s' \in S$, $a \in A$;
$AP$ is a set of atomic propositions;
and $Lab:S \rightarrow 2^{AP}$ is a labelling function, such that $p \in Lab(s)$ iff $p$ is true in $s\in S$.

We define the set of \emph{enabled} actions in $s \in S$ as $A_s=\{a \in A \ | \ \delta_\mathcal{M}(s,a,s')>0 \mbox{ for some } s'\in S \}$.
An infinite \emph{path} through an MDP is a sequence $\smash{\sigma=s_0 \stackrel{a_0}{\rightarrow}s_1\stackrel{a_1}{\rightarrow} \dots}$ where  $\delta_{\mathcal{M}}(s_i,a_i,s_{i+1})>0$ for all $i \in \mathbb{N}$. 
A finite path $\smash{\rho=s_0 \stackrel{a_0}{\rightarrow}s_1\stackrel{a_1}{\rightarrow} ... \stackrel{a_{n-1}}{\rightarrow}s_n}$ is a  prefix  of an infinite  path.
%
%
The choice of action to take at each step of the execution of an MDP $\mathcal{M}$ is made by a \emph{policy},
which can base its decision on the history of $\mathcal{M}$ up to the current state.
Formally, a policy is a function $\pi$ from finite paths of $\mathcal{M}$ to actions in $A$
such that, for any finite path $\sigma$ ending in state $s_n$, $\pi(\sigma) \in A_{s_n}$.
In this work, we will use \emph{memoryless} policies (which only base their choice on the current state)
and \emph{finite-memory} policies (which need to track only a finite set of ``modes'').
%

\subsection{Linear Temporal Logic}

\emph{Linear temporal logic} (LTL) is an extension of propositional logic which allows reasoning about infinite sequences of states. 
We provide a brief overview of LTL and its safe/co-safe fragments here, and direct the reader to~\cite{Pnu81} and~\cite{Kupferman2001}, respectively, for a more complete introduction to these two topics.
LTL formulas $\varphi$ over atomic propositions $AP$ are defined using the following grammar:
\begin{equation*}
\varphi::=\mathit{true} \ | \ p \ | \ \neg \varphi \ | \  \varphi \wedge \varphi \ | \ \ltl_next \varphi \ | \ \varphi \until  \varphi, \textrm{ where } p \in AP.
\end{equation*}
The  $\ltl_next$ operator is read ``next'', meaning that the formula it precedes will be true in the next state.
The $\until$  operator is read ``until'', meaning that its second argument will eventually become true in some state, and the first argument will be continuously true until this point. 
The other propositional connectives can be derived from the ones above in the usual way.
Moreover, other useful LTL operators can be derived from the ones above.
Of particular interest for our work are the ``eventually'' operator $\eventually \varphi$, which requires that $\varphi$ is satisfied in some future state, and the ``always'' operator $\always \varphi$, which requires $\varphi$ to be satisfied in all future states: $\eventually \varphi \equiv \true \until \varphi
$ and $\always \varphi \equiv \neg \eventually \neg \varphi$.

Given an infinite path $\sigma$, we write $\sigma \vDash \varphi$ to denote that  $\sigma$ satisfies formula $\varphi$.
Furthermore, we write $\mathit{Pr}_{\mathcal{M},s}^{\max}(\varphi)$ to denote the maximum probability
(over all policies) of satisfying $\varphi$ from state $s$ in MDP $\mathcal{M}$. 
The semantics of full LTL  is defined over infinite paths. 
However, in this work, we  are interested in specifying behaviours that occur within finite time.
So, we use two well-known subsets of LTL for which properties are meaningful when evaluated over finite paths: \emph{safe} and \emph{co-safe} LTL.
These are based on the notions of \emph{bad prefix} and \emph{good prefix}.
A bad prefix for $\varphi$ is a finite path that cannot be extended in such a way that $\varphi$ is satisfied,
and a good prefix for  $\varphi$ is a finite path that cannot be extended in such a way that $\varphi$ is \emph{not} satisfied.
Safe LTL is defined as the set of LTL formulas for which all non-satisfying infinite paths have a finite bad prefix.
Conversely, co-safe LTL is the set of LTL formulas for which all satisfying infinite paths have a finite good prefix.

For simplicity, we assume a \emph{syntactic restriction} for safe and co-safe LTL.
We assume that all formulas are in positive normal form (negation can only appear next to atomic propositions).
Syntactically safe LTL is the set of formulas for which only the $\always$ and $\ltl_next$ temporal operators occur, and syntactically co-safe LTL is the set of formulas for which only the $\ltl_next$, $\eventually$ and $\until$ temporal operators occur.

For any (co-)safe LTL formula $\varphi$ written over $AP$, we can build a deterministic finite automaton (DFA) $\mathcal{A}_\varphi=\langle Q,\overline{q},Q_F,2^{AP},\delta_{\mathcal{A_\varphi}} \rangle$, where: $Q$ is a finite set of states; $\overline{q} \in Q$ is the initial state; $Q_F \subseteq Q$ is the set of accepting states; $2^{AP}$ is the alphabet; and $\delta_{\mathcal{A}_\varphi}:Q \times 2^{AP} \rightarrow Q$ is a  transition function. 
If $\varphi$ is safe,  $\mathcal{A}_\varphi$ is a DFA that accepts exactly the finite paths
(or, more precisely, the sequences of state labellings from those paths)
that are \emph{not} a bad prefix for $\varphi$.
Conversely, if $\varphi$ is co-safe, $\mathcal{A}_\varphi$ is a DFA that accepts exactly the finite paths that are a good prefix for $\varphi$~\cite{Kupferman2001}.
Typically, one wants to \emph{remain} in an accepting state of safe  DFA, thus never generating a bad prefix, and \emph{reach} an accepting state of co-safe DFA, thus generating a good prefix.

\subsection{Optimal Policies for (Co)-Safe Specifications}

For any (co-)safe LTL specification $\varphi$ and MDP $\mathcal{M}$, we can build a \emph{product} MDP $\mathcal{M}_\varphi=\mathcal{M}\otimes\mathcal{A}_\varphi=\langle S \times Q,\overline{s_\varphi},A,\delta_{\mathcal{M}_\varphi},AP,Lab_\varphi\rangle$.
$\mathcal{M}_\varphi$ behaves like the original MDP $\mathcal{M}$, but is augmented with information about the satisfaction of $\varphi$.
Once a path of $\mathcal{M}_\varphi$ reaches an \emph{accepting state} (i.e., a state of the form $(s,q_F)$ for $q_F\in Q_F$), it is a good prefix for $\varphi$ if $\varphi$ is safe, or a bad prefix if it is co-safe.
We then know that $\varphi$ is satisfied, or not satisfied, respectively.
The construction of the product MDP $\mathcal{M}_\varphi$  is well known (see, e.g.,~\cite{Baier2008})
and is such that it preserves the probabilities of paths from $\mathcal{M}$.
Thus, we can reduce, for example, the problem of finding a policy for $\mathit{Pr}_{\mathcal{M},s}^{\max}(\varphi)$
for a co-safe $\varphi$ to a reachability problem in the product MDP $\mathcal{M}_\varphi$, for which optimal policies can be found using standard techniques such as value iteration~\cite{Put94}.
Such policies are memoryless in $\mathcal{M}_\varphi$, and thus finite-memory in $\mathcal{M}$, with $|Q|$ modes.

\section{Simultaneous Task Allocation and Planning under Uncertainty (STAPU)} \label{sec:stapu}

\subsection{Problem Formulation} \label{sec:formulation}

Let $R=\{r_1,...,r_n\}$ be a set of robots (agents).
The operation of each individual robot $r_i$ as it attempts
to perform tasks is modelled by an MDP $\mathcal{M}_i$.
Probabilities in the MDP may represent either uncertainty in its environment or the possibility of failure.
For the latter, we assume that $\mathcal{M}_i$ has a designated failure state
from which, once reached, the robot cannot execute more tasks.

Consider a \emph{mission} $M=(\Phi, \varphi_{\safe})$ where
$\Phi = \{\varphi_1, ..., \varphi_m\}$ is a set co-safe LTL \emph{task specifications}
and $\varphi_{\safe}$ is a \emph{safety specification}.
We assume the mission to fulfil the two decomposition properties used in~\cite{Schillinger2016}, in particular LTL formulas must be independent, i.e., (non-)satisfaction of one specification must not violate any other specification in the mission; and completion of all mission specification implies the completion of the overall mission (defined as the conjunction of all specifications).

We define a \emph{simultaneous task allocation and planning under uncertainty (STAPU)} problem
as finding a task allocation mapping  $T^*:\Phi \rightarrow R$ such that:

\begin{equation}
\label{eq:stapu}
T^* = \max_{\{T:\Phi \rightarrow R\}} \prod_{i=1}^n \mathit{Pr}_{\M_i}^{\max}(\varphi^{T}_i)
\end{equation}

where the LTL specification $\varphi^{T}_i$ is defined as
the conjunction of tasks for robot $i$ given task allocation $T$:

\begin{equation}
\varphi^{T}_i =  \varphi_{\safe} \wedge  \bigwedge_{\{\varphi \in \Phi \ | \ T(\varphi) = r_i \}} \varphi
\end{equation}

We also need to compute the corresponding optimal policies
$\pi_1,\dots,\pi_n$ for the MDPs $\mathcal{M}_1,\dots,\mathcal{M}_n$.
Since we assume task independence, solving a STAPU problem is effectively finding a joint policy
(i.e. allocation of tasks and the actions performed by each robot)
that maximises the probability of the team achieving the mission.
For now, we assume that a task that is in progress when a robot fails
is never completed; we will see how to deal with this situation in Section~\ref{sec:unfold}.

\subsection{Solution}

In order to solve the problem described above efficiently, we extend the approach proposed in~\cite{Schillinger2016},
which ignores possible physical interactions between robots and exploits the assumption that tasks have no interdependencies and can each be completed by a single robot.
Although tasks will ultimately be executed by robots in parallel, solving a single STAPU problem is done using a sequential model in which we consider each robot independently in turn,
avoiding the construction of the  fully synchronised team model\footnote{In such a model, commonly known as a \emph{multi-agent} MDP~\cite{boutilier1996tark}, both states and actions are considered jointly, which results in an exponential blow-up of the number of states and of the action space.}.

Each independent model is a \emph{local product MDP}, encoding the dynamics of an individual robot,
the definitions of the tasks and the extent to which they have so far been completed.
These models are joined into a \emph{team MDP} through the use of \emph{switch transitions} which represent
changes in the allocation of a task from robot $r_i$ to $r_{i+1}$ (the next robot in the sequential model). 
For the STAPU problem we add switch transitions from every state in robot $r_i$'s model where it completes a task
to every initial state for $r_{i+1}$.
This next robot has an initial state for every possible combination of allocated tasks. 
Considered sequentially, this model allows each robot a choice, on task completion, of whether it or the subsequent robot should tackle the next task. When the model is solved to create a team policy,
the switch transitions result in a policy which creates the optimal task allocation across the team  that optimises Equation~(\ref{eq:stapu}), along with the optimal action choices for each robot, i.e. a solution to the STAPU problem. 
We formalise these steps below.

\vspace*{0.3em}
\subsubsection{Local Product MDPs}\label{ssec:local_mdp}

The first step of the approach is encoding the task definitions into each robot model $\M_i$.
To do so, we build the product MDP $\M_i^M = \M_i \otimes \A_{\varphi_1} \otimes \dots \otimes \A_{\varphi_m} \otimes \A_{\varphi_{\safe}} = \langle S_i^M, \overline{s_i^M}, A_i, \delta_{\mathcal{M}_i^M}, AP, Lab \rangle$,
i.e., we include a separate DFA for each LTL formula making up the mission specification $M$.
The state space of the resulting MDP is
$S_i^M = S_i \times Q_{\varphi_1} \times ... \times Q_{\varphi_n} \times  Q_{\varphi_{\safe}}$,
allowing us to keep track of the state of satisfaction for each part of the mission specification separately.
Finding a policy that maximises the probability of achieving a subset of the mission tasks
can be done by calculating the policy that maximises the probability, in $\M_i^M$,
of reaching the accepting states of the corresponding DFAs whilst remaining in an accepting state of $\A_{\varphi_{\safe}}$.

\vspace*{0.3em}
\subsubsection{Team MDP} \label{sec:team_mdp}
The team MDP $\mathcal{G}$ is the union of the $n$ local product  MDPs $\M_1^M,\dots,\M_n^M$.
Instead of building a fully synchronised multi-agent MDP, we build upon the approach of~\cite{Schillinger2016} and construct a team MDP that represents each robot sequentially.
More precisely, we build the team MDP $\mathcal{G} = \langle S_{\mathcal{G}}, \overline{s_{\mathcal{G}}}, A_{\mathcal{G}}, \delta_{\mathcal{G}},AP_{\mathcal{G}},Lab_{\mathcal{G}}\rangle$ where:
\begin{itemize}
\item $S_{\mathcal{G}}$ keeps track of the robot currently being considered and its current state (within its local product MDP):
\begin{equation*}
\bigcup\nolimits_{i=1}^n \, \{i\}\!\times S_i^M 
\end{equation*}
\item $\overline{s_{\mathcal{G}}} = (1, \overline{s_1^M})$, i.e., we start planning for robot $r_1$, with it in its initial state;
\item $A_{\mathcal{G}} = \{\zeta\} \,\cup\, \bigcup_{i = 1}^n A_i $, i.e., the action space comprises the individual robot actions plus a special \emph{switch transition}~$\zeta$ that is used to make the planning process move from allocating tasks for $r_i$ to allocating tasks for $r_{i+1}$;
\item For $a \in \bigcup_{i = 1}^n A_i $,  the transition function mirrors the corresponding local product transition function:
\begin{equation*}
 \delta_{\mathcal{G}}((i, s_i), a, (i, {s_{i}}')) = \delta_{\M_i^M}( s_i, a,  {s_{i}}')
\end{equation*}
For $a = \zeta$, the transition function updates the system state such that it can start planning for the next robot, i.e.,  $\delta_{\mathcal{G}}((i, s), \zeta, (j,s')) = 1$ if all of the following conditions hold:
\begin{itemize}
\item  $j = 1 + (i \textrm{ mod } n)$, i.e., we connect the robots in a ring topology;
\item The state of all the tasks is preserved and we do not switch during the execution of a task, i.e., we keep all the DFA components of the state the same, and they must correspond to either the initial or the accepting state of the DFA: $s = (s_i, \bold{q})$ and $s' = (s_j, \bold{q})$,
where $\bold{q} = (q_{\varphi_1},...,q_{\varphi_m}, q_{\varphi_{\safe}})$
and each $q_\varphi$ is either the initial state $\overline{q}$ or an accepting state in $Q_F$
from the corresponding DFA $\A_{\varphi}$;
\item $s'$ corresponds to an initial state of robot $j$, i.e.,  $s' = (\overline{s_j}, \bold{q})$.
\end{itemize}
For all other pair of states, $\delta_{\mathcal{G}}((i, s), \zeta, (j,s')) = 0$.
We omit details of the propositions $AP_{\mathcal{G}}$ and labelling function $Lab_{\mathcal{G}}$,
since they are not required here.
\end{itemize}

Note that in the team MDP task allocation and planning are addressed in a sequential fashion, but the generated policies are to be executed in parallel by the team.
This can create ambiguity at execution time.
For example, robot $i+1$'s policy might have different actions corresponding to different possible (probabilistic) executions of robots $1,..., i$'s policies.
However, when execution is starting, robot $i+1$ still does not know how robots $1,..., i$'s policies will evolve.
This means that at the start of execution robot $i+1$  does not know which part of  its policy should be executed.
In fact, this is a source of \emph{partial observability} as robot $i+1$ can only have a \emph{belief} over what the execution of robots $1,..., i$ will be.

In the current paper, we tackle this by imposing
restrictions over the probabilistic nature of the underlying models.
In particular, to avoid the ambiguity described above, we require that the single robot models $\M_i$ are such that the policy that optimises Equation~(\ref{eq:stapu}) contains at most one switch transition per robot. 
Note that, after building the optimal policy, we can easily check for the uniqueness of switch transitions.
However, it is not straightforward to check if arbitrary single robot MDPs fall within this class without solving the team MDP.
This motivates our use of a particular class of single robot MDPs where this uniqueness of switch transitions occurs:  MDPs with a designated \emph{failure state} $s_\bot \in S$ such that,
for all $s\in S,a \in A_s$ and for some $s'\in S$, either $\delta(s,a,s') = 1$ or $\delta(s,a,s')+\delta(s,a,s_\bot)=1$. i.e., where actions either move to a next state $s'$ with probability $x$ or fail with probability $1-x$. 
For such MDPs, the optimal policy for the team MDP has a single switch transition for each robot $i$, corresponding to the state where robot $i$ has executed all the tasks it was allocated with (note however that the robot might fail execution; this will be addressed in the next section).
In our experiments, the single robot MDPs will be instances of this class.

%
%

The solution of a team MDP formed using the aforementioned class of MDPs solves the STAPU problem since we assume that tasks are independent and ignore robot interactions.
More concretely,  the order in which tasks are completed is not relevant, due to task independence and we can plan for each robot ignoring the state of the other robots due to the non-interacting robot assumption.
Given that the team MDP encompasses all possible task allocations and all possible ways  a robot might complete each task, maximising the probability of reaching an accepting state for \emph{all} DFA components is equivalent to  finding a sequence of policies that optimise Equation~(\ref{eq:stapu}) when executed in a parallel fashion.

\section{STAPU with Reallocation} \label{sec:unfold}

In the previous section, when a task fails to be completed by a robot, it is removed from the set achievable by the system. However, this task can be \emph{reallocated} to another member of the team for completion.
In this section we extend the approach described above such that tasks are redistributed among the other team members when a robot fails.
In order to do so, we add new switch transitions which are used to perform this reallocation.

When a failure occurs, robots have already started executing policies obtained from the solution of the STAPU problem.
So, the outcomes of the switch transitions for reallocation must describe in which states the robots might be  \emph{at the point in time when the failure occurred}.
This need for synchronisation across multiple robots breaks the assumptions which allowed us to build a sequential model for the basic STAPU solution. 
 
In order to tackle this issue, while avoiding building a full joint multi-agent MDP, we  build a joint model representing the synchronised evolution of the system \emph{just under the computed STAPU policies}, i.e., we build the synchronised multi-robot policy for the STAPU solution, where, at each state, each robot executes the action corresponding to its policy in a joint fashion.
This model has a set of \emph{reallocation states}, corresponding to situations where some robot has failed (i.e., the probability of that robot achieving more tasks has become $0$).
We then choose one of these reallocation states as the initial state of a new STAPU instance, adding switch transitions to it.

\begin{figure}
\vspace*{0.5em}
\includegraphics[width=\columnwidth]{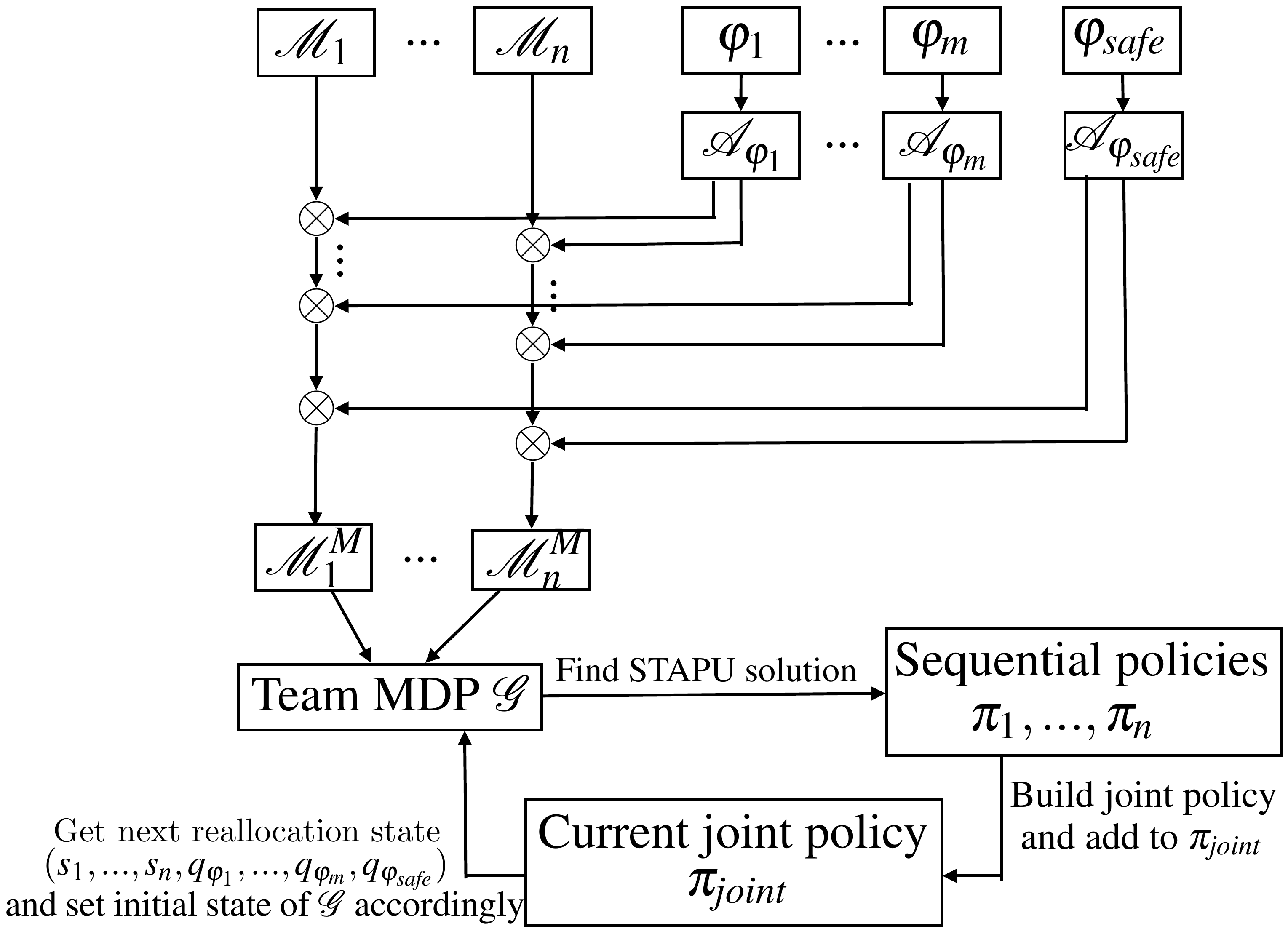} 
\caption{Outline of the overall approach. The mission MDP $\M_i^M$ for each robot $r_i$ is built as the product of the robot MDP and the specification DFAs. Then, we build the team MDP $\G$ and solve a STAPU for the initial state of the robots. The obtained sequential policies are then used to build a synchronised joint policy $\pi_{\joint}$. A reallocation state is chosen from  $\pi_{\joint}$, the initial state of the team MDP, along with its switch transitions, are updated to represent the chosen reallocation state, and a new STAPU is solved. We keep choosing new reallocation states and solving new STAPUs until no more reallocation states exist, or the procedure is interrupted, at which point the current joint policy $\pi_{\joint}$ is returned, along with the associated guarantees on probability of achieving the mission. } \label{fig:diag}
\vspace*{-0.6em}
\end{figure}

More precisely,  let $(s_1,...,s_n,q_{\varphi_1},...,q_{\varphi_m},q_{\varphi_{\safe}})$ be such a reallocation state in the synchronised multi-robot policy, where $s_i$ is the failure state  to be addressed.
We create a new (sequential) team MDP as described in~\ref{sec:team_mdp} except we now consider the initial state to be $\overline{s_{\G}} = (s_i, q_{\varphi_1},...,q_{\varphi_m},q_{\varphi_{\safe}})$ and the switch transitions now point to the current state  of the next robot in the reallocation state $s_{1 + (i \mod n)}$ rather than to the initial state of the next robot $\overline{s_{1 + (i \mod n)}}$.
By solving this new STAPU instance, we effectively find a reallocation policy from the chosen reallocation state.
We can then use the reallocation policy to continue building the synchronised multi-robot policy from $(s_1,...,s_n,q_{\varphi_1},...,q_{\varphi_m},q_{\varphi_{\safe}})$, and then choose a new reallocation state to address.
We choose the next reallocation state to address in decreasing order of reachability probability, i.e., we start by addressing the most probable reallocation states.
When all reallocation states have been addressed, we terminate the procedure.
Note that because we choose the next reallocation state to address in decreasing order of reachability probability the algorithm is anytime, in that it can be terminated at any point and provide a (incomplete) solution, and the longer it runs the more cases it will cover.
Fig.~\ref{fig:diag} shows a summary of the full procedure for STAPU with reallocation.

We finish by noting that this approach can be  extended to also include expected team sum of costs minimisation by defining a cost structure (representing navigation duration for example) for each robot and using nested value iteration (NVI)~\cite{NestedVI} to generate policies.
NVI enforces a strict preference of objectives, by choosing, from all policies that maximise the probability of success, one that minimises the expected cumulative cost.

\section{EVALUATION}
\label{sec:eval}
We implement STAPU with reallocation in the PRISM model checker~\cite{PRISM}, which supports solving MDPs for LTL properties.
%
We evaluate the scalability of the approach with respect to the number of robots and tasks in the model.
 
For our evaluation, we assume homogeneous robots performing simple reachability tasks. 
We create an MDP model of a topological map for one of the environments in the \emph{Patrolling Sim} simulator~\cite{PatrolSim}. The map consists of 30 states. 
In this MDP model, we assume that when navigating to certain states, which we will refer to as \emph{failure points}, the robots can fail (meaning the robot MDP can move to its failure state $s_\bot$) with a certain probability.
This matches the class of models for which our STAPU solution yields single robot policies that can be executed in parallel without ambiguity, as discussed in Section~\ref{sec:stapu}.
The \emph{mission} $M=(\Phi, \varphi_{\safe})$  is specified as $\Phi = \{\varphi_1, ..., \varphi_m\}$ where $\varphi_i = \eventually \ p_i$ and $\varphi_{\safe} = \always \neg \ p$, $\ p_i \in AP$. 

\begin{table}

\caption[caption]{Comparing STAPU with reallocation and VI over the fulll MAMDP model}

\label{tab:sizes}
\begin{center}
   \begin{tabular}{ |l|l|l|l|l|l|l|l|l| }
  \hline
   Tasks & 
  \multicolumn{3}{|c|}{STAPU with reallocation} & 
   \multicolumn{3}{|c|}{VI over MAMDP model}\\
  \hline
  & \multicolumn{2}{|c|}{Model Size} & Time &  \multicolumn{2}{|c|}{Model Size} & Time \\
  
   & $|S_G|$ & $|\delta_G|$ & (s) & $|S_G|$ & $|\delta_G|$ & (s) \\ 
   \hline
  3 & 480 & 1442 & 0.09 & 7200 & 64800 & 1.17 \\
5 & 1920 & 5788 & 0.12 & 28800 & 259200 & 11.96 \\
7 & 7680 & 23226 & 0.51 & 115200 & 1036800  & 195.55 \\ 
9 & 30720 & 93176 & 4.40 & 460800 & 4147200  & 3218.47 \\ 
 
  \hline
\end{tabular}

\end{center}
\vspace*{-0.8em}

\end{table}

Table~\ref{tab:sizes} compares model sizes and computation time for STAPU with an approach based on using value iteration over the full multi-agent MDP (MAMDP) model, which we also implemented in PRISM. Due to the large size of the resulting MAMDP, we limit the number of robots to 2 and the number of failure points to 5 per robot model.
We can see that our approach entails significant gains in both model sizes as well as solution times.
Furthermore, whilst the policy generated from STAPU with reallocation may be sub-optimal, in our experiments with the class of MDP models of topological maps described above,  STAPU with reallocation  achieved the same probability of satisfaction as directly solving the MAMDP, i.e., our approach yielded the optimal solution.
%
 

%

%

\pgfplotsset{every axis/.append style={
                    label style={font=\tiny},
                    tick label style={font=\tiny},
                              }                    }
                              
\begin{figure}
\begin{center}
\vspace*{0.2em}
\begin{tikzpicture}
\begin{groupplot}[
     group style = {group size = 2 by 1,
     },
     width = 0.5\columnwidth,
     set layers,cell picture=true,]

    \nextgroupplot[
  title=(a) Computation Time - 4 Robots,
  title style={font=\tiny,anchor=center,yshift=-0.3em},
  cycle multi list={dashed\nextlist red,blue,green,yellow,magenta\nextlist mark=*},
  xlabel=Tasks,
  xtick=data,
  ylabel= Time (s),
  legend to name={CommonLegend},legend style={legend columns= 7, font=\tiny},
  every axis y label/.style={
  	at={(ticklabel cs:0.5)},anchor=center,rotate=90,font=\tiny},
  	xtick align=inside, 
  	ytick align=inside,
  	yticklabel pos=lower,
  	yticklabel shift={-11.25em},
  	ymajorgrids=true,
  	xmajorgrids=true,
  	  	    every axis x label/.style={
  	at={(ticklabel cs:0.5)},anchor=center,font=\tiny},
  	xlabel shift={0em}
 ]
 \addlegendimage{empty legend}
 \addlegendentry{\hspace{-0.16\columnwidth}Failure Points:}
\addplot table [x=Tasks,y expr=\thisrow{5}/1000]{task_TotalTime_4.dat};
\addlegendentry{$5$}
\addplot table [y expr=\thisrow{10}/1000, x=Tasks]{task_TotalTime_4.dat};
\addlegendentry{$10$}
\addplot table [y expr=\thisrow{15}/1000, x=Tasks]{task_TotalTime_4.dat};
\addlegendentry{$15$}
\addplot table [y expr=\thisrow{20}/1000, x=Tasks]{task_TotalTime_4.dat};
\addlegendentry{$20$}
\addplot table [y expr=\thisrow{25}/1000, x=Tasks]{task_TotalTime_4.dat};
\addlegendentry{$25$}

   \nextgroupplot[
  title=(b) Computation Time - 8 Robots,
   title style={font=\tiny,anchor=center,yshift=-0.3em},
  cycle multi list={dashed\nextlist red,blue,green,yellow,magenta\nextlist mark=*},
  xlabel=Tasks,
  xtick=data,
  ylabel= Time (s),
    every axis y label/.style={
  	at={(ticklabel cs:0.5)},anchor=center,rotate=90,font=\tiny},
  	xtick align=inside, 
  	ytick align=inside,
  	yticklabel pos=lower,
  	yticklabel shift={-1.5em},
  	ymajorgrids=true,
  	xmajorgrids=true,
  	  	    every axis x label/.style={
  	at={(ticklabel cs:0.5)},anchor=center,font=\tiny},
  	xlabel shift={0em}
 ]
\addplot table [x=Tasks,y expr=\thisrow{5}/1000]{task_TotalTime_8.dat};

\addplot table [y expr=\thisrow{10}/1000, x=Tasks]{task_TotalTime_8.dat};

\addplot table [y expr=\thisrow{15}/1000, x=Tasks]{task_TotalTime_8.dat};

\addplot table [y expr=\thisrow{20}/1000, x=Tasks]{task_TotalTime_8.dat};

\addplot table [y expr=\thisrow{25}/1000, x=Tasks]{task_TotalTime_8.dat};

\end{groupplot}

\path (group c1r1.south east) -- node[below=2.5ex]{\ref{CommonLegend}} (group c2r1.south west);

\end{tikzpicture}

\end{center}
\vspace*{-0.8em}
\caption{Total time taken by STAPU with reallocation to generate a complete policy for teams of 4 and 8 robots with varying tasks.}\label{fig:comp_time_etc}
\vspace*{-0.8em}

\end{figure}
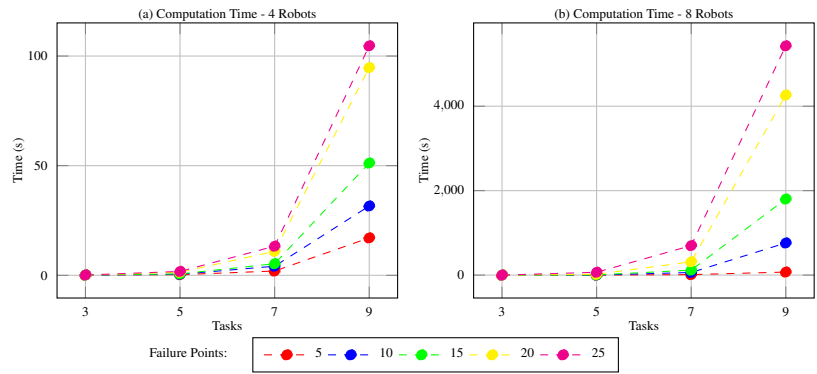 

Fig.~\ref{fig:comp_time_etc} shows that keeping all other factors constant, the total computation time including model building and solving increases exponentially with the number of tasks. This is a consequence of the way the local product is built~(\ref{ssec:local_mdp}), where each new task is added by doing the product of the MDP with the DFA corresponding to that task.

\pgfplotsset{every axis/.append style={
                    label style={font=\tiny},
                    tick label style={font=\tiny},
                              }                    }
                              
\begin{figure}
\begin{center}

\begin{tikzpicture}
\begin{groupplot}[
     group style = {group size = 2 by 1,
     },
     width =0.5\columnwidth,
     set layers,cell picture=true,]
     
  \nextgroupplot[
  title=(a) Number of Reallocations - 4 Robots,
   title style={font=\tiny,anchor=center,yshift=-0.3em},
  cycle multi list={dashed\nextlist red,blue,green,yellow,magenta\nextlist mark=*},
   legend to name={CommonLegend2},legend style={legend columns= 6, font=\tiny},
  xlabel=Failure Points (per robot),
  xtick=data,
  ylabel= Reallocations,
    every axis y label/.style={
  	at={(ticklabel cs:0.5)},anchor=center,rotate=90,font=\tiny},
  	xtick align=inside, 
  	ytick align=inside,
  	yticklabel pos=lower,
  	yticklabel shift={-1.125em},
  	ymajorgrids=true,
  	xmajorgrids=true,
  	  	    every axis x label/.style={
  	at={(ticklabel cs:0.5)},anchor=center,font=\tiny},
  	xlabel shift={0em}
 ]
 \addlegendimage{empty legend}
 \addlegendentry{\hspace{-0.1\columnwidth} Tasks:}
\addplot table [x=FS,y=3]{fs_SE_4.dat};
\addlegendentry{$3$}
\addplot table [y=5, x=FS]{fs_SE_4.dat};
\addlegendentry{$5$}
\addplot table [y=7, x=FS]{fs_SE_4.dat};
\addlegendentry{$7$}
\addplot table [y=9, x=FS]{fs_SE_4.dat};
\addlegendentry{$9$}

   \nextgroupplot[
  title=(b) Number of Reallocations - 8 Robots,
   title style={font=\tiny,anchor=center,yshift=-0.3em},
  cycle multi list={dashed\nextlist red,blue,green,yellow,magenta\nextlist mark=*},
  xlabel=Failure Points (per robot),
  xtick=data,
  ylabel= Reallocations,
    every axis y label/.style={
  	at={(ticklabel cs:0.5)},anchor=center,rotate=90,font=\tiny},
  	xtick align=inside, 
  	ytick align=inside,
  	yticklabel pos=lower,
  	yticklabel shift={-1.5em},
  	ymajorgrids=true,
  	xmajorgrids=true,
  	    every axis x label/.style={
  	at={(ticklabel cs:0.5)},anchor=center,font=\tiny},
  	xlabel shift={0em}
 ]
\addplot table [x=FS,y=3]{fs_SE_8.dat};

\addplot table [y=5, x=FS]{fs_SE_8.dat};

\addplot table [y=7, x=FS]{fs_SE_8.dat};

\addplot table [y=9, x=FS]{fs_SE_8.dat};

\end{groupplot}

\path (group c1r1.south east) -- node[below=2.5ex]{\ref{CommonLegend2}} (group c2r1.south west);

\end{tikzpicture}

\end{center}
\vspace*{-0.8em}
\caption{Number of reallocations performed with respect to the number of failure points per robot model.}\label{fig:realloc}
\vspace*{-0.8em}
\end{figure}
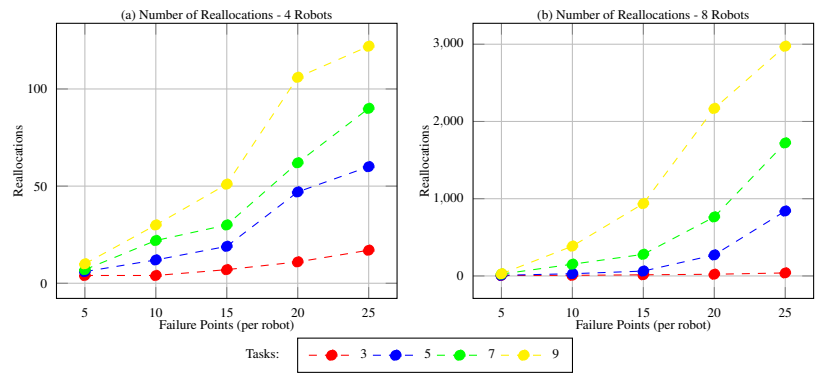   
                             

As the number of failure points grows, more reallocations take place. Fig.~\ref{fig:realloc} shows the number of reallocations for multiple failure points per robot model. More reallocations also increase the computation time. As noted in Section \ref{sec:unfold} an anytime approach can be beneficial in reducing these steps. Since reallocations are performed assuming the initial policy will be executed, a hybrid online-offline version of STAPU is a natural extension of this work.

\section{CONCLUSION}

We have presented an approach to simultaneous task allocation and planning
in multi-robot systems which are operating in uncertain environments and prone to failures, building on techniques for LTL model checking of MDPs.
%
Future work includes handling the policy ambiguity issues yielded by our sequential approach to planning, thus extending the approach presented here to more general MDP models; optimising the policy generation process,
e.g. through the re-use of reallocation states;  incorporating more realistic models of time and robot collisions; and exploiting our anytime approach for online policy execution.



\section*{ACKNOWLEDGMENT}
B. Lacerda and N. Hawes have received funding from UK Research and Innovation and EPSRC through the Robotics and Artificial Intelligence for Nuclear (RAIN) research hub [EP/R026084/1].
D. Parker has received funding from DARPA through the PRINCESS project [FA875016C0045]

\bibliographystyle{IEEEtran}
\bibliography{IEEEabrv,iros}

\end{document}